\documentclass{article}
\usepackage{ijcai25}

\usepackage{times}
\usepackage{soul}
\usepackage{url}
\usepackage[hidelinks]{hyperref}
\usepackage[utf8]{inputenc}
\usepackage[small]{caption}
\usepackage{graphicx}
\usepackage{amsmath}
\usepackage{amsthm}
\usepackage{booktabs}
\usepackage{algorithm}
\usepackage{algorithmic}
\usepackage{multirow} 
\usepackage{xcolor}         
\usepackage[switch]{lineno}
\usepackage{array}
\usepackage{amssymb} 
\usepackage{helvet}  
\usepackage{courier}  
\usepackage{colortbl}

\title{INT: Instance-Specific Negative Mining for Task-Generic \\Promptable Segmentation}

\author{
    Jian Hu, Zixu Cheng, Shaogang Gong
    \affiliations
    Queen Mary University of London
    \emails
    \{jian.hu,zixu.cheng,shaogang.gong\}@qmul.ac.uk
}

\begin{document}

\maketitle

\begin{abstract}
    Task-generic promptable image segmentation aims to
    achieve segmentation of diverse samples under a single task
    description by utilizing only one task-generic prompt.
    Current methods leverage the generalization capabilities
    of Vision-Language Models (VLMs) to infer instance-specific
    prompts from these task-generic prompts in order to guide the segmentation process. 
    However, when VLMs struggle to generalise to some image instances,
    predicting instance-specific prompts becomes poor. 
    To solve this problem, we
    introduce \textbf{I}nstance-specific \textbf{N}egative Mining
    for \textbf{T}ask-Generic Promptable Segmentation (\textbf{INT}). 
    The key idea of INT is to adaptively reduce the influence of
    irrelevant (negative) prior knowledge whilst to increase the use
    the most plausible prior knowledge, selected by negative mining
    with higher contrast, in order to optimise instance-specific prompts generation.
    Specifically, INT consists of two components: (1) instance-specific
    prompt generation, which progressively fliters out incorrect
    information in prompt generation; (2) semantic mask generation,
    which ensures each image instance segmentation matches correctly the
    semantics of the instance-specific prompts. 
    INT is validated on six datasets, including
    camouflaged objects and medical images, demonstrating its effectiveness,
    robustness and scalability. 
\end{abstract}

\section{Introduction}
Task-generic promptable image segmentation leverages a
task-generic prompt to derive instance-specific prompts for segmenting
diverse images given a single task-query. 
Unlike traditional segmentation methods that require per-instance distintive and
exhaustive prompting (labels) of different instances in a dataset, this approach to image
segmentation only requires a single task-generic prompt applicable to all
test samples in a target domain therefore more desirable in practice, e.g. “the polyp” is a task-generic
prompt for all images in a polyp segmentation task. 
However, this approach is also more challenging due to the lack of
an instance-specific prompt (label) in segmenting each image.

\begin{figure}[ht]
   \centering \includegraphics[width=8.5cm]{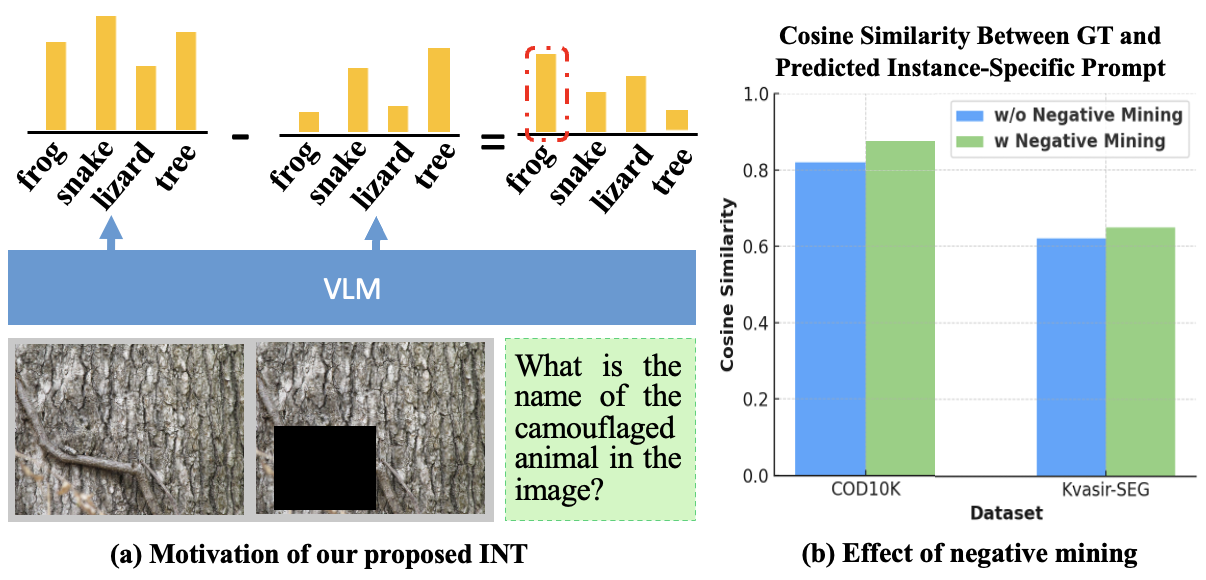}
   \vspace{-15pt}
   \caption{(a) Motivation of INT. When task-related objects
   in the input to the VLM are occluded, the unique features of these
   objects are also obscured, leading to significant changes in the
   corresponding VLM output. In contrast, the features of other
   objects, which are not fully occluded, show only minor changes in
   the VLM output. We leverage this observation to assess the
   correctness of the generated instance-specific prompts without the
   need for ground truth. By progressive negative mining, we iteratively
   correct difficult-to-identify erroneous prompts. 
(b) Evaluation of INT. CLIP semantic similarities are compared between
   the instance-specific prompts INT generated and the ground
   truth. INT's contrastive negative mining mechanism effectively
   corrects erroneous samples, ensuring that the generated
   instance-specific prompts are instance-wise optimised. 
   }\label{fig:motivation}
\vspace{-10pt}
\end{figure}

A task-generic prompt, being both coarse and potentially ambiguous, can lead to poor segmentation when directly applied.
Existing solutions \cite{hu2023relax,liu2023grounding} proposed to use
VLMs to mine information from images, deriving instance-specific
prompts from a generic task prompt to guide SAM in segmenting
task-related objects. 
However, they often struggle with complex images when task-relevant
objects are hard to distinguish visually. This leads to inaccurate
segmentation and poor model performance. 
ProMaC~\cite{hu2024leveraging} tackled this problem by exploring
hallucinations as {\em a priori} knowledge. It masks the objects of
interest (foreground) to induce hallucinations, thereby extracting
per-instance background information relevant to segmenting the
foreground. It helps to reason and optimise iteratively instance-specific prompts. 
Although these methods have had some success, the lack of ground-truth
labels on samples makes it hard to validate if the deduced prompts are
accurate. As a result, once an incorrect prompt is generated, error
propogation is difficult to avoid and may increase over time.
%

Despite the lack of manual annotation, the output changes of the VLMs
show how they link candidate categories to the task. As shown in
Fig.~\ref{fig:motivation}(a), a frog is hidden among branches that
blend with the tree trunk, making it tough to spot. Using the VLM
directly, even though the frog triggers a high activation, the tree
trunk incorrectly scores higher for the 'snake' class, causing a
misclassification of the branch as `snake'. If we cover the likely
area of the camouflaged animal, the identical feature identifying the
`frog' is hidden, but the 'snake' features remain visible on the tree
trunk and bark. This leads to a significant decrease in the `frog'
score, while the `snake' score is barely affected. 
This drastic alteration in the VLM response when key features are occluded provides valuable insight: we can assess the accuracy of prompts by observing changes in VLM responses when parts of an image are hidden. The element with greater variation is more likely to be the correct instance-specific prompt.
Moreover, the difference between a correct category and easily
mistaken classes in VLM output disparity is often small, leading to
incorrect object identification when relying on a single output. In
Fig. ~\ref{fig:motivation}(b), under the premise of including the
groundtruth mask, we randomly drew 2 masks and calculated the VLM
changes before and after covering these masks, multiplying them by
category to implement hard false class negative mining. Compared to
using just one mask for covering, the instance-specific prompt
inferred in this way has a higher CLIP semantic similarity to the
groundtruth class. 
However, previous methods lacked mechanisms to utilize such insights, making it difficult to correct wrong predictions without annotations, leading to error propagation during iterations.
Inspired by the observation in Fig.~\ref{fig:motivation}(b), we leverage changes in VLM outputs as a metric to progressively eliminate the influence of irrelevant categories, effectively addressing this issue.

Specfically, in this work, we propose INT, which aims
to gradually eliminate the impact of incorrect categories through
progressive negative mining, utilizing changes in the VLM's output. It
consists of instance-specific prompt generation and semantic mask
generation. 
For prompt generation, the model splits the input image into different
patches and processes them in parallel. In these patches, objects of
interest might be fully or partially visible. 
This variation induces the model to use its pre-learned knowledge to
predict the existance of objects with their names and locations within a patch.
The names and locations of objects predicted in each patch are then sent to an image inpainting module, which erases the predicted objects and fills the space with the surrounding background.
The outputs of the VLMs are compared before and after the image inpainting, and the predictions with the largest output difference are selected as the instance-specific prompts for this iteration. 
The normalized weights of these differences are multiplied with the
VLM's output in the next iteration. This approach helps iteratively
improve the segmentation of hard-to-distinguish categories and allows
for correction of initially misclassified samples as the iterations
continue. 
For mask generation, GroundingDINO~\cite{liu2023grounding} is deployed
to locate task-related objects in the image. 
The detected bounding boxes and prompts are then processed by SAM and
refined using semantic similarity with CLIP.  
Masks above the similarity threshold are combined to produce the segmentation for this iteration, and the corresponding soft mask is applied to the original image to enhance segmentation in the next iteration. \textbf{Out contributions are as follows:}
(1). We introduce INT, a training-free test-time adaptation approach
that uses progressive negative mining to identify more accurate
instance-specific prompts in the absence of annotations, enabling
task-generic promptable segmentations to be more accurate.
(2). Progressive negative mining identifies hard-to-distinguish error categories by cumulatively multiplying the changes in VLM outputs before and after masking across multiple iterations by category, thereby ensuring the accuracy of the generated instance-specific prompts.
(3). Experiments on six datasets demonstrate the effectiveness of our method.
\section{Related Works}
\noindent\textbf{Vision Language Models} 
(VLMs) are adept at handling tasks in both vision and vision-language modalities. They encompass visual understanding models \cite{krizhevsky2012imagenet,radford2021learning,liu2023improvedllava}, visual generation models \cite{ramesh2021zero,hu2020discriminative}, and general-purpose Interfaces \cite{alayrac2022flamingo}.
Visual understanding models develop robust visual representations that serve as the backbone for various computer vision downstream tasks. Thanks to extensive image-text datasets, visual generation models are equipped to tackle a range of visual tasks, such as segmentation and object detection in images or videos \cite{kirillov2023segment,openai2024gpt4o}, based on multimodal inputs.
Although current models have succeeded in addressing many downstream visual tasks, they struggle with specific challenges such as medical image segmentation due to the scarcity of relevant data, making it difficult to achieve high performance in these demanding areas.
Our INT introduces negative mining, employing VLMs to iteratively refine the correct prompts to enhance segmentation performance in these challenging tasks.

\noindent\textbf{Promptable Segmentation} involves segmenting objects with user-provided inputs, such as points, boxes, or scribbles. Models like SAM~\cite{kirillov2023segment}, AV-SAM~\cite{mo2023av}, GroundingSAM~\cite{liu2023grounding}, and SEEM~\cite{zou2023segment} extend this to multimodal inputs, including video and audio. However, these approaches often depend on manual prompts, which are prone to ambiguity and subjectivity, and are typically effective only for specific tasks. GenSAM~\cite{hu2023relax} introduces a manual-free setting, using a task-generic prompt for instance-specific segmentation across images without additional user inputs. It leverages VLMs to infer object names as prompts for SAM, but its lack of spatial information can result in inaccurate predictions in complex scenes.
ProMaC~\cite{hu2024leveraging} introduced hallucination as a type of prior knowledge to leverage task-relevant information as much as possible to aid in the generation of more accurate instance-specific prompts.
However, since there are no ground truths, how to effectively evaluate the quality of instance-specific prompts and exclude the effects of erroneous prompts becomes an unresolved problem.

\noindent\textbf{Prompt Engineering} is a developing field that focuses on creating and refining prompts to improve the effectiveness of large language models (LLMs) for a variety of tasks, encompassing both language and vision modifications.
In the language domain, zero-shot prompting~\cite{wei2021finetuned} is employed to leverage LLMs' generalization capabilities for new tasks, although it can lead to inaccurate outcomes. Recent advancements include chain-of-thought prompting~\cite{wei2022chain} and graph prompting~\cite{liu2023graphprompt}, which enhance complex reasoning skills. Other strategies such as generated knowledge prompting~\cite{liu2021generated} and self-consistency~\cite{wang2022self} have also been implemented to boost prediction accuracy.
For vision-related tasks, prompt tuning is the primary method, using vision-driven~\cite{jia2022visual,zhou2023zegclip}, language-driven~\cite{zhou2022learning,ma2023understanding}, and vision-language driven approaches~\cite{zang2022unified,xing2022class} to create vision prompts that enhance model performance.
Despite some successes in multimodal prompt engineering works~\cite{zhang2023multimodal,hu2023relax,hu2024leveraging} in generating instance-specific prompts, these methods struggle to accurately identify correct prompts without annotations. Our INT approach effectively addresses this challenge by leveraging progressive negative mining.

\begin{figure*}[ht]
   \centering \includegraphics[width=18cm]{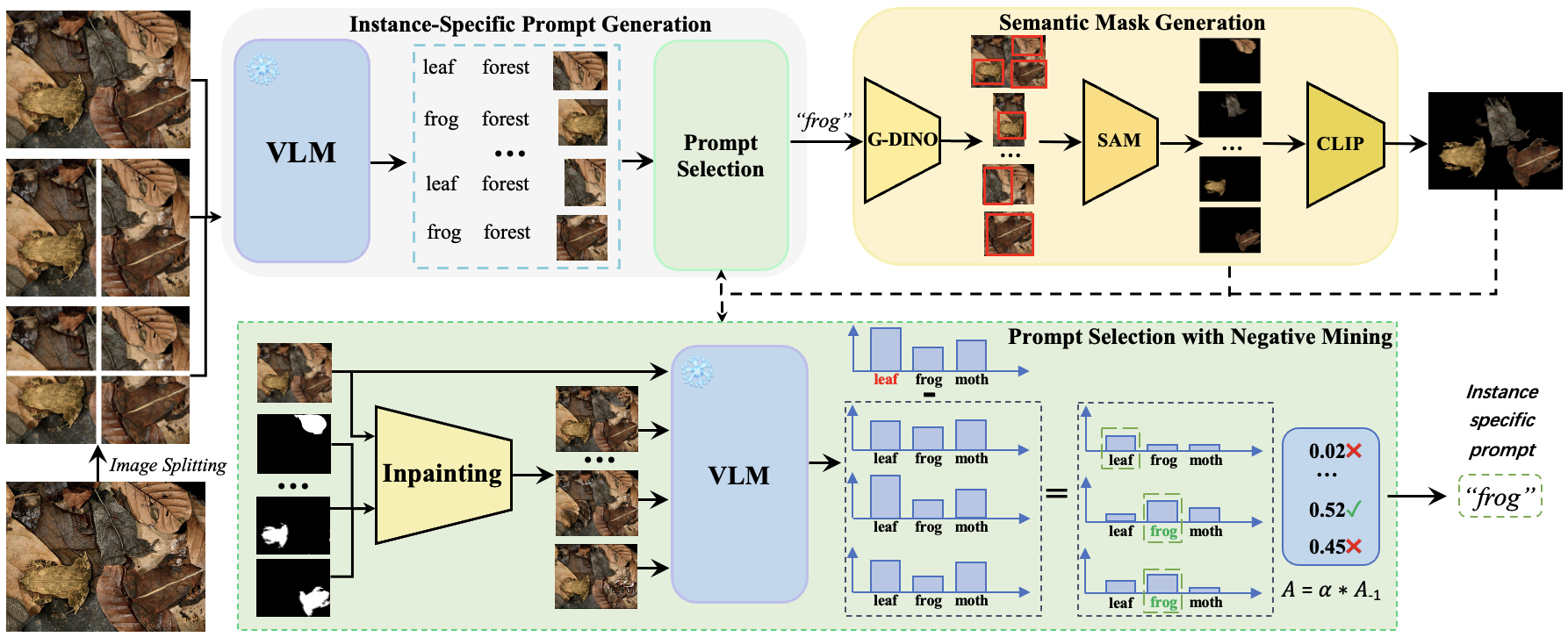}
   \caption{INT consists of two main components: instance-specific prompt generation and semantic mask generation. Initially, the former uses VLMs to generate candidate instance-specific prompts. A prompt selection module then selects the prompt with the highest VLM output contrast, refined through progressive negative mining. This selected prompt is passed to the semantic mask generation module, which employs GroundingDINO to ensure that all task-relevant samples in the image are collected as comprehensively as possible. Simultaneously, SAM and CLIP work together to ensure that the generated masks are semantically aligned with the task. 
   }\label{fig:framework}
\vspace{-10pt}
\end{figure*}

\section{Methodology}
We present INT, a training-free cycle-generation method that segments multiple unknown classes of objects with only a single task-generic prompt. 
This innovative approach leverages negative mining to iteratively reduce the impact of potential erroneous predictions derived from task-generic prompts in an unlabelled setting. 
Specifically, for an image \(X \in \mathbb{R}^{H \times W \times 3}\) from a test set, INT utilizes a task-generic prompt \(P_g\) to generate a final segmentation mask \(M \in \mathbb{R}^{H \times W}\). This eliminates the need for separate supervision for each image, streamlining the process across datasets in the same task category.
The prompt generator identifies multiple candidates for instance-specific prompts, which are evaluated by comparing the VLM outputs before and after image removal. The differences in the outputs are normalized and iteratively weighted, influencing the selection of instance-specific prompts in subsequent iterations. In each iteration, the candidate with the largest difference is chosen as the instance-specific prompt. This selected prompt guides the segmentation process, which further refines the generation of improved instance-specific prompts in the next iteration.
%

\subsection{Instance-Specific Prompt Generation}
\noindent\textbf{Prompt Generation Using VLMs.}
For more accurate segmentation, the prompt utilizes VLMs to transform a generic prompt $P_g$ into instance-specific prompts for each image. 
Specifically, given an image $X$ and a query $P$, the VLM with parameters $\theta$ generates a response that captures task-relevant information. The image $X$ provides essential visual context, aiding the model in formulating a relevant response $y$, which is derived auto-regressively from a probability distribution conditioned on $P$, $X$, and the previous tokens:
\begin{equation}
\label{eq:VLM}
    y_t \sim p_{\theta}(y_t \mid X, P, y_{<t}) \propto \exp (\text{logit}_{\theta}(y_t \mid X, P, y_{<t}))
\end{equation}
where \(y_t\) denotes the token at time \(t\) and \(y_{<t}\) represents the sequence of tokens generated up to time \(t-1\).
Despite the advanced capabilities of VLMs, task-relevant objects may blend into their background due to factors like texture, color, or position, making the instance-specific prompts they generate prone to inaccuracies.
Inaccurate instance-specific prompts can directly lead to incorrect segmentation. However, without any labels, assessing whether these prompts meet the task requirements becomes challenging.
However, VLM predictions are often driven by the unique features of the objects they detect. When such features are obscured, the model's predictions for the corresponding category significantly drop (see Fig. 2). Therefore, if task-related objects can be accurately located and effectively removed, comparing the VLM outputs before and after removal provides a reliable method for evaluating the quality of the instance-specific prompts.

\noindent\textbf{Hallucination-driven Candidates Generation.}
To generate accurate instance-specific prompts, it is essential to identify as many candidates as possible with a reasonable level of confidence to ensure that the correct prompt is not overlooked.
Inspired by ProMaC~\cite{hu2024leveraging}, we divide the input image into patches of varying scales by cutting it horizontally, vertically, both, or leaving it uncut. Each patch is then processed individually by the VLM to generate preliminary instance-specific prompts.
The visibility of task-relevant objects differs across patches, encouraging the VLM to leverage its prior knowledge to infer potential objects and their locations. This process helps identify candidate bounding boxes and object names by linking the visual data in each patch with the task context.
The VLM processes each patch as follows:
\begin{small}
\begin{equation}
\label{eq: naive_bbox_object}
    B^{k} = \operatorname{VLM}(X^k, C^k, P_{B}), \quad
    A^{k}_\text{fore}, A^{k}_\text{back} = \operatorname{VLM}(X^k, C^k, P_{A}),
\end{equation}
\end{small}
where $C^k$ is the caption for the $k$-th image patch $X^k$, and $P_g$ is the task-generic prompt.
For bounding box predictions, the prompt $P_B$ guides the VLM: ``{\em{This image pertains to the $P_g$ detection task, output the bounding box of the $P_g$.}}'' This instructs the VLM to predict the bounding boxes $B^k$ for objects related to the task within the patch.
For object naming, the prompt $P_A$ states: ``{\em{Output the name of the $P_g$ and its environment in one word.}}'' This directs the VLM to predict the names of the task-related objects $A_{\text{fore}}^k$ and their backgrounds $A_{\text{back}}^k$ from each patch.
Object names $A^k_{\text{fore}}$ and bounding boxes $B^k$, collected from different patches, are compiled into candidate lists $\mathcal{A}_i$ and $\mathcal{B}_i$, where $i$ denotes the iteration.

\noindent\textbf{Prompts Selection with Negative Mining.}
After generating the candidate lists, we evaluate which candidates are most likely correct. When a prediction is accurate, removing the corresponding object causes a significant change in the VLM's output for that category. Building on the previous section, where we ensured the ground truth was included, we now mask candidate-indicated areas and measure the change in the VLM's output scores.
To achieve this, inspired by contrastive decoding, we compare the VLM softmax output of the original unprocessed image patch $X_k$ with that of the masked patch $X_k^{'}$ for each patch k. The category with the largest difference in output is selected as the final prediction for the corresponding patch. This approach ensures that the most significant and task-relevant object is accurately identified and used as the instance-specific prompt as follows:
\begin{align}
    D(y^k_i) =& \max(\text{softmax}(\text{logit}_\theta(y_t \mid X_k, P, y_{<t})) \notag\\&- \text{softmax}(\text{logit}_\theta(y_t \mid X^{'}_k, P, y_{<t}))),
\end{align}
here, $X^{'}_k$ is processed by the image inpainting module, which uses the predicted mask $m_{i}^{k}$ from the last iteration in Sec.~\ref{sec:Semantic Mask Generator} as the inpainting mask $\text{IM}_i^{k}$ to guide the modification of the patch $X^{'}_k$. To ensure that $X^{'}_k$ remains free of task-related objects $A_{i}^\text{fore}$, we employ a negative prompt $P_n$: \emph{``\( A_{i}^\text{fore} \) is not a \( P_g \).''} For seamless Integration with the existing background \( A_{i}^\text{back} \), a positive prompt \( P_p \) is used: \emph{``\( A_{i}^\text{back} \), high quality, detailed, and well-Integrated with the original image.''} 
$X^{'}_k$ is formalized as:
\begin{equation}
    X^{'}_k = F_{in}(X_k, \text{IM}_i^{k}, P_p, P_n),
\end{equation}
where $F_{in}$ represents the inpainting module, implemented using Stable Diffusion.
Using this method, we construct a set $D(y) = \{D(y^k_i)\}_{k=1}^{K}$, where $K$ represents the number of patch. The accuracy of the predicted category is directly related to the magnitude of change in unique information about the correct category before and after masking. By evaluating these changes, we identify the most reasonable foreground $A_i^u$ to serve as the instance-specific prompt for the $i$-th iteration.
Specifically, we compare the variations in $D(y^k_i)$ across patches and select the candidate with the largest change in output as the final prediction:
\begin{equation}
    A_i^u = \text{argmax}_{k} D(y^k_i), \quad k = 1, 2, \ldots, K,
\end{equation}
This approach leverages the variations in softmax outputs to refine the instance-specific prompt $A_i^u$, ensuring they are closely aligned with task-relevant information and enhancing the overall prediction accuracy.

\noindent\textbf{Progressive Negative Mining}
While this method effectively identifies correct categories when task-related objects are easily distinguishable, it struggles with more ambiguous samples. In such cases, incorrect categories may occasionally exhibit large score differences before and after masking during certain iterations but show little to no change in others. In contrast, the correct category consistently demonstrates stable and significant differences across all iterations. 
Building on this observation, we design a progressive negative mining technique to iteratively refine $A_i^u$. The idea is to accumulate and reinforce consistent patterns by iteratively multiplying scores from each iteration, reducing the influence of unstable changes caused by incorrect categories. 
To normalize the differences from each iteration for comparability, we use:
\begin{equation}
    D_{\text{norm}}(y^k_i) = \frac{D(y^k_i)}{\sum_{k=1}^K D(y^k_i)},
\end{equation}
where $D_{\text{norm}}(y^k_i)$ represents the normalized difference for category $y^k$ at iteration $i$, and K is the total number of patches.
We then iteratively update the differences for the next iteration by applying cumulative multiplication:
\begin{equation}
    D(y^k_{i+1}) = D(y^k_{i+1}) \cdot D_{\text{norm}}(y^k_i),
\end{equation}
where $D(y^k_{i+1})$ is the updated difference for category $y^k$ in iteration $i+1$, incorporating the influence of the current iteration's normalized differences.
By iteratively applying negative mining, we amplify consistent patterns in the score differences associated with the correct category while suppressing the sporadic and unstable changes caused by incorrect categories, effectively enhancing the model’s ability to accurately refine $A_i^u$ over successive iterations.
\subsection{Semantic Mask Generation}
\label{sec:Semantic Mask Generator}
After obtaining the instance-specific prompt $A_i^u$, we aim to produce a mask that accurately delineates task-related objects without overlooking any targets. To achieve this, we first process $A_i^u$ through Grounding DINO~\cite{liu2023grounding} to gather all potential bounding boxes across various patches:
\begin{equation}
    B_i^k = \text{GroundingDINO}(X^k, A_i^u)
\end{equation}
Subsequently, both \(B_i^k\) and \(A_i^u\) are input into SAM:
\begin{equation}
    m_i^k = \text{SAM}(\text{Spatial CLIP}(A_i^u, X_i^k), B_i^k, X_i^k),
\end{equation}
where Spatial CLIP~\cite{hu2023relax} maps the text prompt $A_i^u$ to regions within the image $X_i$ as the visual prompts. These visual prompts, along with the corresponding bounding box $B_i^k$ are fed into SAM during the $i$-th iteration to generate the mask $m_i^k$.
These processed masks, along with the generated instance-specific text prompts $A_i^u$, are subsequently input into CLIP to evaluate semantic similarity.
\begin{equation}
    s(m_i^k) = \text{CLIP}(m_i^k \odot X_i, A_i^u),
\end{equation}
where the operation $\odot$ retains only those parts of $X_i$ covered by the predicted mask.
$s(m_i^k)$ quantifies the similarity between the masked image and \(A_i^u\). Similarity scores from various patches are represented as \(S_i = [s(m_i^1), s(m_i^2), \ldots, s(m_i^k)]\). After normalizing these elements within \(S_i\), a normalized \(s(m_i^k)\) closer to 1 indicates a higher semantic alignment of \(m_i^k\) with the instance-specific text prompt \(A_i^u\). The weighted sum of the normalized \(s(m_i^k)\) and \(m_i^k\) is then computed as follows:
\begin{equation}
    M_i = \sum_{k=1}^{K}(s(m_i^k) * m_i^k),
\end{equation}
where \(M_i\) is the resultant mask from the \(i\)-th iteration of \(X\). This mask, generated using SAM’s capabilities, ensures highly detailed mask production. Concurrently, this mask semantic alignment process guarantees that the output mask is consistent with the task’s semantic requirements, overcoming the limitations of SAM's mask prediction.

The mask is then applied to the original image as a weighting factor to generate the next iteration image \(X_i\) for segmentation. This helps to exclude irrelevant regions and reduce Interference during segmentation:
\begin{align}
    X_{i+1} = w \cdot (X_i \odot M_i) + (1-w) \cdot X_i,
\end{align}
where \(w\) is a hyperparameter set to 0.3.
The mask generated in the last iteration serves to guide the prompt generator in the subsequent iteration, focusing on potential task-related regions, mitigating the impact of irrelevant hallucinations, and yielding more precise instance-specific prompts.
These prompts, in return, aid the mask generator in producing improved masks. Through iterative cycles of prompt and mask generation, both elements enhance significantly.
Ultimately, masks from different iterations are averaged, and the mask closest to this mean is selected as the final output:
\begin{equation}
\mathrm{i}^* = \arg\min_{i}\left(\left| M_i - \frac{\sum_{i}{ (M_{1}, \ldots, M_{\mathbf{I}}})}{i_{\text{result}}} \right|\right).
\end{equation}
Here, \(\mathbf{I}\) represents the number of adaptation epochs, and \(M_{\mathrm{i}^*}\) is the definitive mask for image \(X\).
\begin{table*} 
    \centering
    \setlength{\tabcolsep}{4pt}
    \caption{Results on Camouflaged Object Detection (COD) under different settings. Best are in \textbf{bold}.}
    \vspace{-9pt}
\label{tab:results}
 \renewcommand{\arraystretch}{0.9}
 \resizebox{1.0\textwidth}{!}{
\begin{tabular}{c|c|cccc|cccc|cccc}
\cline{1-14}
{\multirow{3}{*}{Methods}} & \multicolumn{13}{c}{Camouflaged Object Detection}\\\cline{2-14}
& {\multirow{2}{*}{Venue}} &\multicolumn{4}{c|}{
CHAMELEON~\shortcite{skurowski2018animal}}&\multicolumn{4}{c|}{
CAMO~\shortcite{le2019anabranch}}&\multicolumn{4}{c}{
COD10K~\shortcite{fan2021concealed}}\\\cline{3-14}
& & \small{$M\downarrow$} & \small{$F_{\beta}\uparrow$} & \small{$E_{\phi}\uparrow$} &  \small{$S_{\alpha}\uparrow$}& \small{$M\downarrow$} & \small{$F_{\beta}\uparrow$} & \small{$E_{\phi}\uparrow$} &  \small{$S_{\alpha}\uparrow$} & \small{$M\downarrow$} & \small{$F_{\beta}\uparrow$} & \small{$E_{\phi}\uparrow$} &  \small{$S_{\alpha}\uparrow$} \\
\hline
\multicolumn{14}{c}{Scribble Supervision Setting} \\
\hline
WSSA\cite{zhang2020weakly} & \small{CVPR20} & 0.067 & 0.692 & 0.860 & 0.782 & 0.118 & 0.615 & 0.786& 0.696 & 0.071 & 0.536 & 0.770 & 0.684 \\
SCWS\cite{yu2021structure} & \small{AAAI21} & 0.053 & 0.758 & 0.881 & 0.792 & 0.102 & 0.658 & 0.795 & 0.713 & 0.055 & 0.602 & 0.805 & 0.710\\
TEL\cite{zhang2020weakly} & \small{CVPR22} & 0.073 & 0.708 & 0.827 & 0.785 & 0.104 & 0.681 & 0.797 & 0.717 & 0.057 & 0.633 & 0.826 & 0.724 \\
SCOD\cite{he2023weakly} & \small{AAAI23} & \textbf{0.046} & \textbf{0.791} & \textbf{0.897} & 0.818 & \textbf{0.092} & 0.709 & 0.815 & 0.735 & 0.049 & 0.637 & 0.832 & 0.733 \\
SAM-S\cite{kirillov2023segment} & \small{ICCV23} & 0.076 & 0.729 & 0.820 & 0.650 & 0.105 & 0.682 & 0.774 &0.731 & 0.046 & 0.695 & 0.828 & 0.772 \\
WS-SAM\cite{he2023weakly1}& \small{NeurlPS23} & \textbf{0.046} & 0.777 & \textbf{0.897} & \textbf{0.824} & \textbf{0.092} & \textbf{0.742} & \textbf{0.818} & \textbf{0.759} & \textbf{0.038} & \textbf{0.719} & \textbf{0.878} & \textbf{0.803}\\
\hline
\multicolumn{14}{c}{point Supervision Setting} \\
\cline{1-14}
WSSA\cite{zhang2020weakly} & \small{CVPR20} & 0.105 & 0.660 & 0.712 & 0.711 & 0.148 & 0.607 & 0.652 & 0.649 & 0.087 & 0.509 & 0.733 & 0.642 \\
SCWS\cite{yu2021structure} & \small{AAAI21} & 0.097 & 0.684 & 0.739 & 0.714 & 0.142 & 0.624 & 0.672 & 0.687 & 0.082 & 0.593 & 0.777 & 0.738\\
TEL\cite{zhang2020weakly} & \small{CVPR22} & 0.094 & 0.712 & 0.751 & 0.746 & 0.133 & 0.662 & 0.674 & 0.645 & 0.063 & 0.623 & 0.803 & 0.727 \\
SCOD\cite{he2023weakly} & \small{AAAI23} & 0.092 & 0.688 & 0.746 & 0.725 & 0.137 & 0.629 & 0.688 & 0.663 & 0.060 & 0.607 & 0.802 & 0.711 \\
SAM\cite{kirillov2023segment} & \small{ICCV23} & 0.207 & 0.595 & 0.647 & 0.635 & 0.160 & 0.597 & 0.639 & 0.643 & 0.093 & 0.673 & 0.737 & 0.730 \\
SAM-P\cite{kirillov2023segment} & \small{ICCV23} & 0.101 & 0.696 & 0.745 & 0.697 & 0.123 & 0.649 & 0.693 & 0.677 & 0.069 & 0.694 & 0.796 & 0.765\\
WS-SAM\cite{he2023weakly1} & \small{NeurlPS23} & \textbf{0.056} & \textbf{0.767} & \textbf{0.868} & \textbf{0.805} & \textbf{0.102} & \textbf{0.703} & \textbf{0.757} & \textbf{0.718} & \textbf{0.039} & \textbf{0.698} & \textbf{0.856} & \textbf{0.790}\\\hline
\multicolumn{14}{c}{Task-Generic Prompt Setting} \\ 
\cline{1-14}

{\small CLIP\_Surgey+SAM} & \small{Arxiv23} & 0.147 & 0.606 & 0.741 & 0.689 & 0.189 & 0.520 & 0.692 & 0.612 & 0.173 & 0.488 & 0.698 & 0.629	  \\
{\small GPT4V+SAM} \cite{openai2024gpt4v,kirillov2023segment} & \small{Arxiv23} &  0.180 & 0.557 & 0.710 & 0.637& 0.206& 0.466& 0.666 & 0.573 & 0.187 & 0.448 &0.672 & 0.601\\
{\small LLaVA1.5+SAM} \cite{liu2023visual,kirillov2023segment} & \small{NeurlPS23} & 0.168& 0.561 & 0.718 & 0.666& 0.314 & 0.401 & 0.585 & 0.501 & 0.170 & 0.530 & 0.728 & 0.662\\
X-Decoder~\cite{zou2023generalized}  & {\small{CVPR23}} &0.124& 0.654 & 0.748 & 0.716 & 0.104& 0.628 & 0.745 & 0.709 & 0.171 & 0.556 & 0.705 & 0.652\\
SEEM~\cite{zou2023segment} & {\small{NeurIPS23}} & 0.094&  0.011 & 0.307 & 0.454 & 0.192 & 0.023 & 0.315 & 0.404 & 0.143 & 0.001 & 0.280 & 0.425\\
GroundingSAM~\cite{kirillov2023segment,liu2023grounding} & {\small{ICCV23}} & 0.122 & 0.662 & 0.776 & 0.744 & 0.157 & 0.656 & 0.753 & 0.707 & 0.085 & 0.670 & 0.813& 0.764\\
GenSAM~\cite{hu2023relax} & \small{AAAI24} & 0.073 & 0.696 & 0.806 & 0.774 & 0.106 & 0.669 & 0.798 & 0.729 & 0.058 & 0.695 & 0.843 & 0.783\\
ProMaC\cite{hu2024leveraging} & \small{NeurIPS24} &  0.044 & 0.790 & 0.899 & 0.833 & 0.090	& 0.725 & 0.846 & 0.767 & 0.042 & 0.716 & 0.876 & 0.805\\
\rowcolor{purple!10}INT & {\small{Ours}}& \textbf{0.039} & \textbf{0.801} & \textbf{0.906} & \textbf{0.842} & \textbf{0.086}	& \textbf{0.734} & \textbf{0.853} & \textbf{0.772} & \textbf{0.037} & \textbf{0.722} & \textbf{0.883} & \textbf{0.808}\\ 
\hline
\end{tabular}
}
\end{table*}

\begin{table*}[ht]
\setlength{\tabcolsep}{5pt}
    \caption{Results for Medical Image Segmentation (MIS) under task-generic prompt setting.}
    \vspace{-9pt}
    \label{tab:results_2m}
    \centering 
     \renewcommand{\arraystretch}{0.9}
  \resizebox{1.0\textwidth}{!}{
\begin{tabular}{c|c|cccc|cccc|cccc}
\hline
\multicolumn{1}{c|}{\multirow{3}{*}{Methods}} &{\multirow{3}{*}{Venue}} & \multicolumn{8}{c}{Polyp Image   Segmentation} &\multicolumn{4}{|c}{Skin Lesion Segmentation}  \\ \cline{3-14} &  &\multicolumn{4}{c|}{CVC-ColonDB~\shortcite{tajbakhsh2015automated}} & \multicolumn{4}{c|}{Kvasir~\shortcite{jha2020kvasir}}      & \multicolumn{4}{c}{ISIC~\shortcite{codella2019skin}}       \\ \cline{3-14}
& &\small{$M\downarrow$} & \small{$F_{\beta}\uparrow$} & \small{$E_{\phi}\uparrow$} &  \small{$S_{\alpha}\uparrow$}
&\small{$M\downarrow$} & \small{$F_{\beta}\uparrow$} & \small{$E_{\phi}\uparrow$} &  \small{$S_{\alpha}\uparrow$}
&\small{$M\downarrow$} & \small{$F_{\beta}\uparrow$} & \small{$E_{\phi}\uparrow$} &  \small{$S_{\alpha}\uparrow$}\\\hline
{\small GPT4V+SAM} \cite{openai2024gpt4v,kirillov2023segment}& {\small{Arxiv23}}& 0.578 & 0.051 &    0.246 & 0.242& 0.614 & 0.128 & 0.236 & 0.253 & 0.514 & 0.387 & 0.366 & 0.334 \\
{\small LLaVA1.5+SAM} \cite{liu2023visual,kirillov2023segment} & {\small{NeruIPS23}}& 0.491 & 0.194 & 0.355 & 0.357 & 0.479 & 0.293 & 0.400 & 0.403 & 0.369	& 0.473	& 0.497	& 0.477
\\
X-Decoder~\cite{zou2023generalized} &  {\small{CVPR23}}& 0.462 & 0.095 & 0.327 & 0.331 & 0.449 & 0.202 & 0.371 & 0.384 & 0.338 & 0.315 & 0.127 & 0.407  \\
SEEM~\cite{zou2023segment} & {\small{NeruIPS23}}& 0.570 & 0.085 & 0.280 & 0.284 & 0.520 & 0.215 & 0.339 & 0.367 & 0.362 & 0.250 & 0.002 & 0.280 \\
GroundingSAM~\cite{kirillov2023segment,liu2023grounding} & {\small{ICCV23}}&    0.711    &   0.071    &   0.195    &    0.206   &   0.387    &  0.353    &    0.521   &    0.468  & 0.301 & 0.348 & 0.247 & 0.533\\ 
GenSAM~\cite{hu2023relax}& {\small{AAAI24}} & 0.244 & 0.059	& 0.494	&0.379 & 0.172 &	0.210 &	0.619 &	0.487 & 0.171 & 0.699 &	0.744 & 0.678 \\
ProMaC~\cite{hu2024leveraging} & {\small{NeurIPS24}}& 0.176	 &  0.243 & 0.583 &0.530 & 0.166	& 0.394	 & 0.726 & 0.573 & 0.160 & 0.728 & 0.766 & 0.703\\ 
\rowcolor{purple!10}INT & {\small{Ours}}& \textbf{0.172}	 &  \textbf{0.250} & \textbf{0.589} &\textbf{0.537} & \textbf{0.161}	& \textbf{0.401}	 & \textbf{0.732} & \textbf{0.5739} & \textbf{0.152} & \textbf{0.733} & \textbf{0.771} & \textbf{0.708}\\ 
\hline
\end{tabular}}
\end{table*}

\section{Experiments}
\label{sec:experiment}

\begin{figure*}[ht]
   \centering   \includegraphics[width=18.5cm]{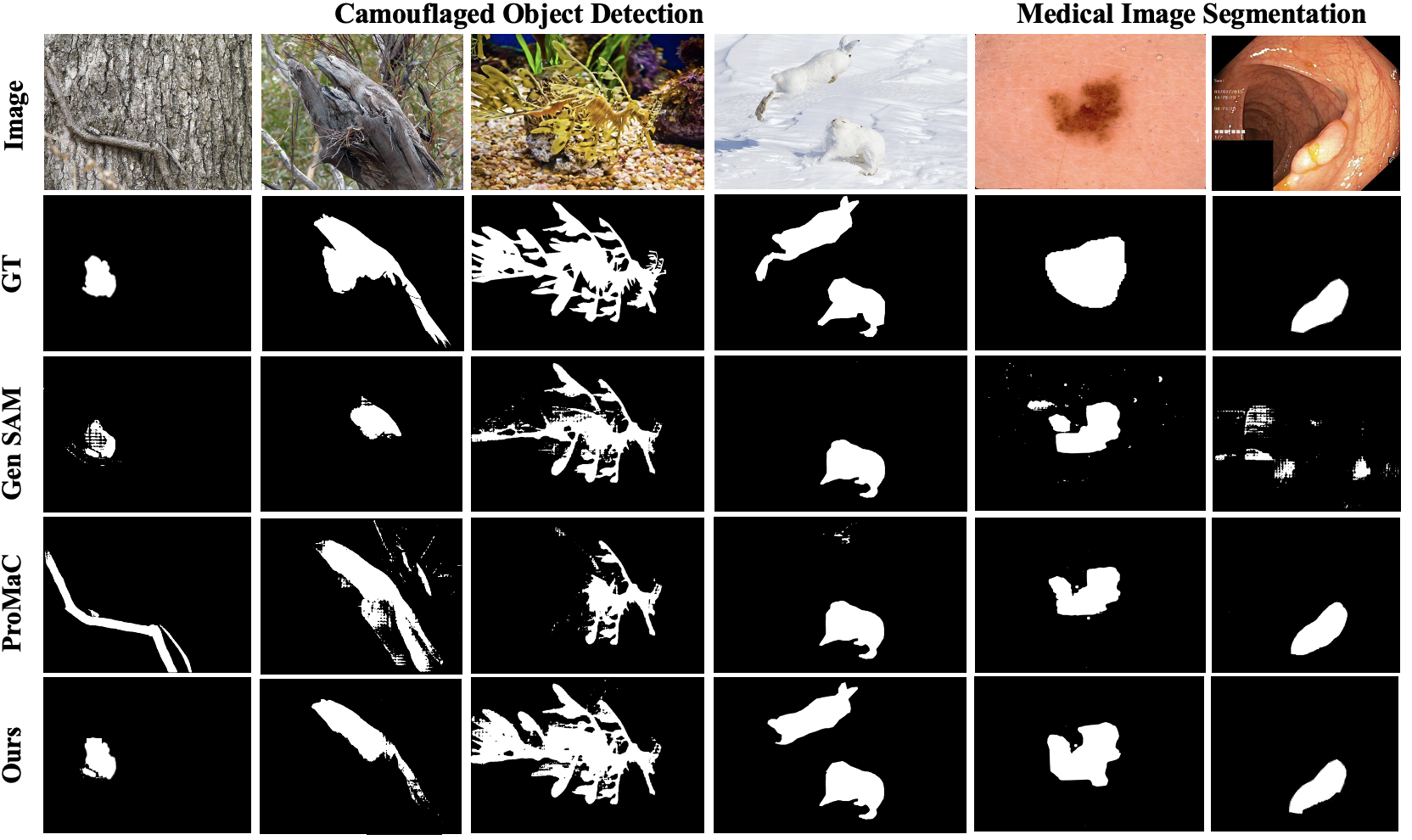}
   \vspace{-15pt}
   \caption{Visualization of various segmentation methods among various segmentation tasks.
}\label{fig:visualization}
\vspace{-6pt}
\end{figure*}

\begin{table*}[tb!]
\caption{{Ablation study on CHAMELEON dataset.}}
  \vspace{-9pt}
\resizebox{1.0\textwidth}{!}{
\footnotesize
  \begin{tabular}
  { @{\hspace{-10pt}}
  c @{\hspace{0pt}}  c@{\hspace{0pt}}
  c@{\hspace{0pt}}}
    { (a) Number of iteration $\mathbf{I}$.\label{table:itertion}}&
     { (b) Image preprocess strategy.\label{table:scale}}&
     {(c) module ablation study. \label{table:Postprocessing}}\\
    {
    \footnotesize
    \setlength{\tabcolsep}{5pt}
    \renewcommand{\arraystretch}{0.68}
        \begin{tabular}{c|cccc} 
        \hline
        $\mathbf{I}$ & {$M\downarrow$} & {$F_{\beta}\uparrow$} & {$E_{\phi}\uparrow$} &  {$S_{\alpha}\uparrow$}\\ \hline
        1  &  0.079 & 0.692 & 0.861 & 0.794 \\
        2   &0.053 & 0.748 & 0.897 & 0.816  \\
        3   & 0.050	& 0.752 & 0.902	& 0.823\\
        4  & 0.045  & 0.792 &  0.903  & 0.829 \\
        \cellcolor{purple!10}5  & \cellcolor{purple!10}0.039 & \cellcolor{purple!10}\textbf{0.801} & \cellcolor{purple!10}\textbf{0.906} & \cellcolor{purple!10}0.842 \\ 
        6 &\textbf{0.038} & 0.800 & \textbf{0.906} & \textbf{0.844} \\\hline
    \end{tabular}}&
    {\setlength{\tabcolsep}{5pt}
    \renewcommand{\arraystretch}{0.97}
        \begin{tabular}{c|cccc} 
        \hline
        Scale      &  {$M\downarrow$} & {$F_{\beta}\uparrow$} & {$E_{\phi}\uparrow$} &  {$S_{\alpha}\uparrow$} \\ \hline
        Original   & 0.071 & 0.541 & 0.758 & 0.668   \\
        Havel   & 0.059	& 0.582	& 0.780 & 0.692  \\
        Quarters & 0.062 & 0.482 & 0.685 &  0.601\\
        Original+Havel   & 0.043 & 0.793 & 0.891 & 0.835   \\
        \cellcolor{purple!10}Original  +Havel+Quarters  & \cellcolor{purple!10}\textbf{0.039} & \cellcolor{purple!10}\textbf{0.801} & \cellcolor{purple!10}\textbf{0.906} & \cellcolor{purple!10}\textbf{0.842} \\ \hline
    \end{tabular}} &
    {\setlength{\tabcolsep}{2pt}
    \renewcommand{\arraystretch}{0.65}
    \begin{tabular}{cccc|cccc}
            \hline
            \multicolumn{4}{c|}{Method's Variants} & \multicolumn{4}{c}{CHAMELEON~\shortcite{skurowski2018animal}}  \\
            \hline
            HCG & PSNM & PNM & SMG &  $M\downarrow$ & $F_{\beta}\uparrow$ & $E_{\phi}\uparrow$ & $S_{\alpha}\uparrow$  \\
            \hline
            & \checkmark & \checkmark & \checkmark & 0.071 & 0.541 & 0.758 & 0.668 \\
            \checkmark & & \checkmark & \checkmark & 0.083 & 0.703& 0.811 & 0.746  \\
            \checkmark & \checkmark & & \checkmark & 0.060 & 0.732 & 0.836 & 0.768  \\
            \checkmark & \checkmark & \checkmark & & 0.053 & 0.772 & 0.895 & 0.818  \\
            \rowcolor{purple!10}
            \checkmark & \checkmark & \checkmark & \checkmark & \textbf{0.039} & \textbf{0.801} & \textbf{0.906} & \textbf{0.842}  \\
            \hline
        \end{tabular}} 
     
  \end{tabular}}
\end{table*}

\subsection{Experimental Setup} 

\noindent\textbf{Baselines.} Our study evaluates the ProMaC model's effectiveness in handling complex segmentation challenges such as Camouflaged Object Detection (COD), Medical Image Segmentation (MIS), and Transparent Object Detection (TOD), where traditional SAM models often fall short \cite{ji2023segment}.
In our COD assessments, ProMaC is benchmarked against various weakly supervised segmentation methods \cite{kirillov2023segment,zhang2020weakly,yu2021structure,zhang2020weakly,he2023weakly,he2023weakly,hu2019multi,hu2022learning,he2023camouflaged,hu2020unsupervised,zhang2021domain}. We employ two levels of supervision for comparison: scribble supervision, where the main structures of foreground and background are delineated during training, and point supervision, where distinct points are annotated for both.
For task-generic prompt settings, we introduce a demanding scenario by relying solely on a task description as a generic prompt for segmentation. Here,INT incorporates LLaVA1.5 \cite{liu2023visual} with SAM \cite{kirillov2023segment}.
Further experimentation in MIS and PIS tasks aims to validate our method's superiority using task-generic prompts versus traditional techniques. We test combinations like GPT4V+SAM and LLaVA1.5+SAM to highlight the limitations of current VLM models in this context. Our INT is also evaluated with leading state-of-the-art promptable segmentation methods to underline its effectiveness.
Our results are the average of three trials.

\noindent\textbf{Metric.} To assess performance in the first three tasks, we employ the following metrics: Mean Absolute Error (M), adaptive F-measure ($F_{\beta}$) \cite{margolin2014evaluate}, mean E-measure ($E_{\phi}$) \cite{fan2021cognitive}, and structure measure ($S_{\alpha}$) \cite{fan2017structure}. A lower M or higher $F_{\beta}$, $E_{\phi}$ and $S_{\alpha}$
indicate superior performance. 

\noindent\textbf{PyTorch Implementation Details.} For the VLM models, we employ LLaVA-1.5-13B for evaluation. For image processing, we use the CS-ViT-B/16 model pre-trained with CLIP, and for image inpainting module, we deploy stable-diffusion-2-inpainting. The task-generic prompts for the COD task are specified as "camouflaged animal." The MIS task includes two sub-tasks: polyp image segmentation and skin lesion segmentation, each prompted by "polyp" and "skin lesion" respectively. All tasks undergo training-free test-time adaptation, iterating for four epochs, except for the polyp image segmentation task, which extends to six epochs. We utilize the ViT-H/16 model for promptable segmentation methods. Our experiments are conducted on a single NVIDIA A100 GPU, with further details provided in the appendix.

\subsection{Results and Analysis}
\noindent\textbf{Results on COD Task.} The COD tasks are designed to detect animals camouflaged within complex environments. We tested INT on three benchmark datasets: CHAMELEON \cite{skurowski2018animal}, CAMO \cite{le2019anabranch}, and COD10K \cite{fan2021concealed}. 
The CHAMELEON dataset includes 76 images collected from the Internet specifically for testing purposes. The CAMO dataset contains 1,250 images, divided into 1,000 training images and 250 testing images. The COD10K dataset comprises 3,040 training samples and 2,026 testing samples in total.
As indicated in Table \ref{tab:results}, we compared INT against other
methods that apply different supervision levels. Generally, methods
with scribble supervision outperformed those with point
supervision. Notably, our INT, utilizing only a single generic task
prompt, outshines all point-supervised methods and scribble-supervised
methods on all three datasets in terms of all the metrics. This
highlights INT's effectiveness. Moreover, INT consistently
surpasses SAM, SAM-P, SAM-S, and CLIP Surgery+SAM, demonstrating that
the enhancements provided by INT are more than leveraging the 
segmentation capability of SAM.  

\noindent\textbf{Results on MIS Tasks.} The MIS task involves identifying pathological tissues in medical images. We utilized datasets such as CVC-ColonDB \cite{tajbakhsh2015automated} and Kvasir \cite{jha2020kvasir} for polyp image segmentation, and ISIC \cite{codella2019skin} for skin lesion segmentation. Comparisons from Table \ref{tab:results_2m} show that we conducted experiments in the task-generic promptable segmentation setting. Since most VLMs have not been specifically trained on medical images, directly applying a VLM to medical tasks results in significantly lower performance compared to natural image tasks. In contrast, our proposed INT, through extensive candidate sample inference and negative mining, effectively explores task-relevant information and progressively filters out incorrect samples. This ensures the accuracy of the generated instance-specific prompts, greatly enhancing INT’s segmentation performance on the MIS task.

\noindent\textbf{Parameter Analysis.} Tab. \ref{table:Postprocessing}(a) investigates how various model metrics evolve over a prolonged number of iterations. all the metrics stabilize after just 5 iterations. Although the results continue to change, the variations remain slight. This indicates that our method converges rapidly and remains stable. It also underscores the importance of using iteration-based termination criteria to establish early stopping conditions.
Tab. \ref{table:Postprocessing}(b) examines the effects of different image processing strategies. "Original" refers to the unmodified image, "Halve" splits the image horizontally or vertically into two parts, and "Quarters" divides it into four smaller patches. Testing results indicate that combining "Original", "Halve" and "Quarters" achieves the best balance between global and local information, avoiding excessive fragmentation.

\noindent\textbf{Module Analysis.} As shown in Tab. \ref{table:Postprocessing}(c), we perform an ablation study on the COD and MIS tasks to assess the effects of different modules.
"HCG" refers to hallucination-driven candidate generation, "PSNM" stands for prompt selection with negative mining, "PNM" is progressive negative mining, and "SMG" refers to the semantic mask generator.
The first row shows that replacing HCG with just a single original image leads to reduced performance, underscoring the importance of using hallucinations to extract task-relevant information.
In the second row, replacing prompt selection with negative mining using the VLM inference result yields worse performance than the full model, highlighting the significance of proper prompt selection.
Removing progressive negative mining results in a significant drop in performance, indicating that the instance-specific prompts derived from the initial iteration may contain errors, and our approach effectively corrects these mistakes.
The comparison between the last two rows emphasizes the importance of aligning the mask with task semantics. The consistently positive results across tasks confirm the robustness and effectiveness of our approach.

\noindent\textbf{Visualization.} Fig. \ref{fig:visualization} visually compares our method, INT, with other approaches across two tasks.
GenSAM performs well with clear objects but struggles in complex backgrounds.
ProMaC produces solid segmentation results across various tasks, but it sometimes misidentifies instance-specific prompts in complex scenes, leading to segmentation results that are unrelated to the task (e.g., the first column in Fig. \ref{fig:visualization}).
In contrast, our INT introduces a negative mining strategy that not only explores potential candidates to extract task-relevant information from the image for better segmentation but also corrects misidentified instance-specific prompts from early iterations. This approach effectively improves performance. Additionally, our method can segment multiple task-related samples within the same image, something that previous methods could not achieve.
It demonstrated the effectiveness of our approach.

\section{Conclusion}
In this work, we introduced Instance-specific Negative Mining for
Promptable Segmentation (INT) for task-generic promptable image
segmentation. INT leverages the difference in
VLM outputs before and after masking as a metric for progressive (iterative)
negative mining. By employing progressive negative mining, INT
predicts more accurate instance-specific prompts from a single coarse task-generic
prompt. This allows for effective segmentation of different targets
within the same task across various images, even in the absence of
annotations. Experiments conducted on six diverse datasets demonstrate
the effectiveness of our INT. 
\bibliographystyle{named}
\bibliography{ijcai25}

\begin{thebibliography}{}

\bibitem[\protect\citeauthoryear{Alayrac \bgroup \em et al.\egroup }{2022}]{alayrac2022flamingo}
Jean-Baptiste Alayrac, Jeff Donahue, Pauline Luc, Antoine Miech, Iain Barr, Yana Hasson, Karel Lenc, Arthur Mensch, Katherine Millican, Malcolm Reynolds, et~al.
\newblock Flamingo: a visual language model for few-shot learning.
\newblock {\em Advances in Neural Information Processing Systems}, 35:23716--23736, 2022.

\bibitem[\protect\citeauthoryear{Codella \bgroup \em et al.\egroup }{2019}]{codella2019skin}
Noel Codella, Veronica Rotemberg, Philipp Tschandl, M~Emre Celebi, Stephen Dusza, David Gutman, Brian Helba, Aadi Kalloo, Konstantinos Liopyris, Michael Marchetti, et~al.
\newblock Skin lesion analysis toward melanoma detection 2018: A challenge hosted by the international skin imaging collaboration (isic).
\newblock {\em arXiv preprint arXiv:1902.03368}, 2019.

\bibitem[\protect\citeauthoryear{Fan \bgroup \em et al.\egroup }{2017}]{fan2017structure}
Deng-Ping Fan, Ming-Ming Cheng, Yun Liu, Tao Li, and Ali Borji.
\newblock Structure-measure: A new way to evaluate foreground maps.
\newblock In {\em Proceedings of the IEEE international conference on computer vision}, pages 4548--4557, 2017.

\bibitem[\protect\citeauthoryear{Fan \bgroup \em et al.\egroup }{2021a}]{fan2021concealed}
Deng-Ping Fan, Ge-Peng Ji, Ming-Ming Cheng, and Ling Shao.
\newblock Concealed object detection.
\newblock {\em IEEE transactions on pattern analysis and machine intelligence}, 44(10):6024--6042, 2021.

\bibitem[\protect\citeauthoryear{Fan \bgroup \em et al.\egroup }{2021b}]{fan2021cognitive}
Deng-Ping Fan, Ge-Peng Ji, Xuebin Qin, and Ming-Ming Cheng.
\newblock Cognitive vision inspired object segmentation metric and loss function.
\newblock {\em Scientia Sinica Informationis}, 6(6), 2021.

\bibitem[\protect\citeauthoryear{He \bgroup \em et al.\egroup }{2023a}]{he2023camouflaged}
Chunming He, Kai Li, Yachao Zhang, Longxiang Tang, Yulun Zhang, Zhenhua Guo, and Xiu Li.
\newblock Camouflaged object detection with feature decomposition and edge reconstruction.
\newblock In {\em Proceedings of the IEEE/CVF conference on computer vision and pattern recognition}, pages 22046--22055, 2023.

\bibitem[\protect\citeauthoryear{He \bgroup \em et al.\egroup }{2023b}]{he2023weakly1}
Chunming He, Kai Li, Yachao Zhang, Guoxia Xu, Longxiang Tang, Yulun Zhang, Zhenhua Guo, and Xiu Li.
\newblock Weakly-supervised concealed object segmentation with sam-based pseudo labeling and multi-scale feature grouping.
\newblock {\em arXiv preprint arXiv:2305.11003}, 2023.

\bibitem[\protect\citeauthoryear{He \bgroup \em et al.\egroup }{2023c}]{he2023weakly}
Ruozhen He, Qihua Dong, Jiaying Lin, and Rynson~WH Lau.
\newblock Weakly-supervised camouflaged object detection with scribble annotations.
\newblock In {\em Proceedings of the AAAI Conference on Artificial Intelligence}, volume~37, pages 781--789, 2023.

\bibitem[\protect\citeauthoryear{Hu \bgroup \em et al.\egroup }{2019}]{hu2019multi}
Jian Hu, Hongya Tuo, Chao Wang, Lingfeng Qiao, Haowen Zhong, and Zhongliang Jing.
\newblock Multi-weight partial domain adaptation.
\newblock In {\em BMVC}, page~5, 2019.

\bibitem[\protect\citeauthoryear{Hu \bgroup \em et al.\egroup }{2020a}]{hu2020discriminative}
Jian Hu, Hongya Tuo, Chao Wang, Lingfeng Qiao, Haowen Zhong, Junchi Yan, Zhongliang Jing, and Henry Leung.
\newblock Discriminative partial domain adversarial network.
\newblock In {\em Computer Vision--ECCV 2020: 16th European Conference, Glasgow, UK, August 23--28, 2020, Proceedings, Part XXVII 16}, pages 632--648. Springer, 2020.

\bibitem[\protect\citeauthoryear{Hu \bgroup \em et al.\egroup }{2020b}]{hu2020unsupervised}
Jian Hu, Hongya Tuo, Chao Wang, Haowen Zhong, Han Pan, and Zhongliang Jing.
\newblock Unsupervised satellite image classification based on partial transfer learning.
\newblock {\em Aerospace Systems}, 3:21--28, 2020.

\bibitem[\protect\citeauthoryear{Hu \bgroup \em et al.\egroup }{2022}]{hu2022learning}
Jian Hu, Haowen Zhong, Fei Yang, Shaogang Gong, Guile Wu, and Junchi Yan.
\newblock Learning unbiased transferability for domain adaptation by uncertainty modeling.
\newblock In {\em European Conference on Computer Vision}, pages 223--241. Springer, 2022.

\bibitem[\protect\citeauthoryear{Hu \bgroup \em et al.\egroup }{2024a}]{hu2023relax}
Jian Hu, Jiayi Lin, Shaogang Gong, and Weitong Cai.
\newblock Relax image-specific prompt requirement in sam: A single generic prompt for segmenting camouflaged objects.
\newblock In {\em Proceedings of the AAAI Conference on Artificial Intelligence}, volume~38, pages 12511--12518, 2024.

\bibitem[\protect\citeauthoryear{Hu \bgroup \em et al.\egroup }{2024b}]{hu2024leveraging}
Jian Hu, Jiayi Lin, Junchi Yan, and Shaogang Gong.
\newblock Leveraging hallucinations to reduce manual prompt dependency in promptable segmentation.
\newblock {\em arXiv preprint arXiv:2408.15205}, 2024.

\bibitem[\protect\citeauthoryear{Jha \bgroup \em et al.\egroup }{2020}]{jha2020kvasir}
Debesh Jha, Pia~H Smedsrud, Michael~A Riegler, P{\aa}l Halvorsen, Thomas de~Lange, Dag Johansen, and H{\aa}vard~D Johansen.
\newblock Kvasir-seg: A segmented polyp dataset.
\newblock In {\em MultiMedia Modeling: 26th International Conference, MMM 2020, Daejeon, South Korea, January 5--8, 2020, Proceedings, Part II 26}, pages 451--462. Springer, 2020.

\bibitem[\protect\citeauthoryear{Ji \bgroup \em et al.\egroup }{2023}]{ji2023segment}
Wei Ji, Jingjing Li, Qi~Bi, Wenbo Li, and Li~Cheng.
\newblock Segment anything is not always perfect: An investigation of sam on different real-world applications.
\newblock {\em arXiv preprint arXiv:2304.05750}, 2023.

\bibitem[\protect\citeauthoryear{Jia \bgroup \em et al.\egroup }{2022}]{jia2022visual}
Menglin Jia, Luming Tang, Bor-Chun Chen, Claire Cardie, Serge Belongie, Bharath Hariharan, and Ser-Nam Lim.
\newblock Visual prompt tuning.
\newblock In {\em European Conference on Computer Vision}, pages 709--727. Springer, 2022.

\bibitem[\protect\citeauthoryear{Kirillov \bgroup \em et al.\egroup }{2023}]{kirillov2023segment}
Alexander Kirillov, Eric Mintun, Nikhila Ravi, Hanzi Mao, Chloe Rolland, Laura Gustafson, Tete Xiao, Spencer Whitehead, Alexander~C Berg, Wan-Yen Lo, et~al.
\newblock Segment anything.
\newblock {\em arXiv preprint arXiv:2304.02643}, 2023.

\bibitem[\protect\citeauthoryear{Krizhevsky \bgroup \em et al.\egroup }{2012}]{krizhevsky2012imagenet}
Alex Krizhevsky, Ilya Sutskever, and Geoffrey~E Hinton.
\newblock Imagenet classification with deep convolutional neural networks.
\newblock {\em Advances in neural information processing systems}, 25, 2012.

\bibitem[\protect\citeauthoryear{Le \bgroup \em et al.\egroup }{2019}]{le2019anabranch}
Trung-Nghia Le, Tam~V Nguyen, Zhongliang Nie, Minh-Triet Tran, and Akihiro Sugimoto.
\newblock Anabranch network for camouflaged object segmentation.
\newblock {\em Computer vision and image understanding}, 184:45--56, 2019.

\bibitem[\protect\citeauthoryear{Liu \bgroup \em et al.\egroup }{2021}]{liu2021generated}
Jiacheng Liu, Alisa Liu, Ximing Lu, Sean Welleck, Peter West, Ronan~Le Bras, Yejin Choi, and Hannaneh Hajishirzi.
\newblock Generated knowledge prompting for commonsense reasoning.
\newblock {\em arXiv preprint arXiv:2110.08387}, 2021.

\bibitem[\protect\citeauthoryear{Liu \bgroup \em et al.\egroup }{2023a}]{liu2023improvedllava}
Haotian Liu, Chunyuan Li, Yuheng Li, and Yong~Jae Lee.
\newblock Improved baselines with visual instruction tuning, 2023.

\bibitem[\protect\citeauthoryear{Liu \bgroup \em et al.\egroup }{2023b}]{liu2023visual}
Haotian Liu, Chunyuan Li, Qingyang Wu, and Yong~Jae Lee.
\newblock Visual instruction tuning.
\newblock {\em arXiv preprint arXiv:2304.08485}, 2023.

\bibitem[\protect\citeauthoryear{Liu \bgroup \em et al.\egroup }{2023c}]{liu2023grounding}
Shilong Liu, Zhaoyang Zeng, Tianhe Ren, Feng Li, Hao Zhang, Jie Yang, Chunyuan Li, Jianwei Yang, Hang Su, et~al.
\newblock Grounding dino: Marrying dino with grounded pre-training for open-set object detection.
\newblock {\em arXiv preprint arXiv:2303.05499}, 2023.

\bibitem[\protect\citeauthoryear{Liu \bgroup \em et al.\egroup }{2023d}]{liu2023graphprompt}
Zemin Liu, Xingtong Yu, Yuan Fang, and Xinming Zhang.
\newblock Graphprompt: Unifying pre-training and downstream tasks for graph neural networks.
\newblock In {\em Proceedings of the ACM Web Conference 2023}, pages 417--428, 2023.

\bibitem[\protect\citeauthoryear{Ma \bgroup \em et al.\egroup }{2023}]{ma2023understanding}
Chengcheng Ma, Yang Liu, Jiankang Deng, Lingxi Xie, Weiming Dong, and Changsheng Xu.
\newblock Understanding and mitigating overfitting in prompt tuning for vision-language models.
\newblock {\em IEEE Transactions on Circuits and Systems for Video Technology}, 2023.

\bibitem[\protect\citeauthoryear{Margolin \bgroup \em et al.\egroup }{2014}]{margolin2014evaluate}
Ran Margolin, Lihi Zelnik-Manor, and Ayellet Tal.
\newblock How to evaluate foreground maps?
\newblock In {\em Proceedings of the IEEE conference on computer vision and pattern recognition}, pages 248--255, 2014.

\bibitem[\protect\citeauthoryear{Mo and Tian}{2023}]{mo2023av}
Shentong Mo and Yapeng Tian.
\newblock Av-sam: Segment anything model meets audio-visual localization and segmentation.
\newblock {\em arXiv:2305.01836}, 2023.

\bibitem[\protect\citeauthoryear{OpenAI}{2024a}]{openai2024gpt4v}
OpenAI.
\newblock Gpt-4v: Enhancing gpt-4 for visual processing.
\newblock 2024.
\newblock Accessed: 2024-05-20.

\bibitem[\protect\citeauthoryear{OpenAI}{2024b}]{openai2024gpt4o}
OpenAI.
\newblock Hello gpt-4o.
\newblock 2024.
\newblock Accessed: 2024-05-20.

\bibitem[\protect\citeauthoryear{Radford \bgroup \em et al.\egroup }{2021}]{radford2021learning}
Alec Radford, Jong~Wook Kim, Chris Hallacy, Aditya Ramesh, Gabriel Goh, Sandhini Agarwal, Girish Sastry, Amanda Askell, Pamela Mishkin, Jack Clark, et~al.
\newblock Learning transferable visual models from natural language supervision.
\newblock In {\em International conference on machine learning}, pages 8748--8763. PMLR, 2021.

\bibitem[\protect\citeauthoryear{Ramesh \bgroup \em et al.\egroup }{2021}]{ramesh2021zero}
Aditya Ramesh, Mikhail Pavlov, Gabriel Goh, Scott Gray, Chelsea Voss, Alec Radford, Mark Chen, and Ilya Sutskever.
\newblock Zero-shot text-to-image generation.
\newblock In {\em International Conference on Machine Learning}, pages 8821--8831. PMLR, 2021.

\bibitem[\protect\citeauthoryear{Skurowski \bgroup \em et al.\egroup }{2018}]{skurowski2018animal}
Przemys{\l}aw Skurowski, Hassan Abdulameer, J~B{\l}aszczyk, Tomasz Depta, Adam Kornacki, and P~Kozie{\l}.
\newblock Animal camouflage analysis: Chameleon database.
\newblock {\em Unpublished manuscript}, 2(6):7, 2018.

\bibitem[\protect\citeauthoryear{Tajbakhsh \bgroup \em et al.\egroup }{2015}]{tajbakhsh2015automated}
Nima Tajbakhsh, Suryakanth~R Gurudu, and Jianming Liang.
\newblock Automated polyp detection in colonoscopy videos using shape and context information.
\newblock {\em IEEE transactions on medical imaging}, 35(2):630--644, 2015.

\bibitem[\protect\citeauthoryear{Wang \bgroup \em et al.\egroup }{2022}]{wang2022self}
Xuezhi Wang, Jason Wei, Dale Schuurmans, Quoc Le, Ed~Chi, Sharan Narang, Aakanksha Chowdhery, and Denny Zhou.
\newblock Self-consistency improves chain of thought reasoning in language models.
\newblock {\em arXiv preprint arXiv:2203.11171}, 2022.

\bibitem[\protect\citeauthoryear{Wei \bgroup \em et al.\egroup }{2021}]{wei2021finetuned}
Jason Wei, Maarten Bosma, Vincent~Y Zhao, Kelvin Guu, Adams~Wei Yu, Brian Lester, Nan Du, Andrew~M Dai, and Quoc~V Le.
\newblock Finetuned language models are zero-shot learners.
\newblock {\em arXiv preprint arXiv:2109.01652}, 2021.

\bibitem[\protect\citeauthoryear{Wei \bgroup \em et al.\egroup }{2022}]{wei2022chain}
Jason Wei, Xuezhi Wang, Dale Schuurmans, Maarten Bosma, Fei Xia, Ed~Chi, Quoc~V Le, Denny Zhou, et~al.
\newblock Chain-of-thought prompting elicits reasoning in large language models.
\newblock {\em Advances in Neural Information Processing Systems}, 35:24824--24837, 2022.

\bibitem[\protect\citeauthoryear{Xing \bgroup \em et al.\egroup }{2022}]{xing2022class}
Yinghui Xing, Qirui Wu, De~Cheng, Shizhou Zhang, Guoqiang Liang, and Yanning Zhang.
\newblock Class-aware visual prompt tuning for vision-language pre-trained model.
\newblock {\em arXiv preprint arXiv:2208.08340}, 2022.

\bibitem[\protect\citeauthoryear{Yu \bgroup \em et al.\egroup }{2021}]{yu2021structure}
Siyue Yu, Bingfeng Zhang, Jimin Xiao, and Eng~Gee Lim.
\newblock Structure-consistent weakly supervised salient object detection with local saliency coherence.
\newblock In {\em Proceedings of the AAAI conference on artificial intelligence}, volume~35, pages 3234--3242, 2021.

\bibitem[\protect\citeauthoryear{Zang \bgroup \em et al.\egroup }{2022}]{zang2022unified}
Yuhang Zang, Wei Li, Kaiyang Zhou, Chen Huang, and Chen~Change Loy.
\newblock Unified vision and language prompt learning.
\newblock {\em arXiv preprint arXiv:2210.07225}, 2022.

\bibitem[\protect\citeauthoryear{Zhang \bgroup \em et al.\egroup }{2020}]{zhang2020weakly}
Jing Zhang, Xin Yu, Aixuan Li, Peipei Song, Bowen Liu, and Yuchao Dai.
\newblock Weakly-supervised salient object detection via scribble annotations.
\newblock In {\em Proceedings of the IEEE/CVF conference on computer vision and pattern recognition}, pages 12546--12555, 2020.

\bibitem[\protect\citeauthoryear{Zhang \bgroup \em et al.\egroup }{2021}]{zhang2021domain}
Shizhao Zhang, Hongya Tuo, Jian Hu, and Zhongliang Jing.
\newblock Domain adaptive yolo for one-stage cross-domain detection.
\newblock In {\em Asian conference on machine learning}, pages 785--797. PMLR, 2021.

\bibitem[\protect\citeauthoryear{Zhang \bgroup \em et al.\egroup }{2023}]{zhang2023multimodal}
Zhuosheng Zhang, Aston Zhang, Mu~Li, Hai Zhao, George Karypis, and Alex Smola.
\newblock Multimodal chain-of-thought reasoning in language models.
\newblock {\em arXiv preprint arXiv:2302.00923}, 2023.

\bibitem[\protect\citeauthoryear{Zhou \bgroup \em et al.\egroup }{2022}]{zhou2022learning}
Kaiyang Zhou, Jingkang Yang, Chen~Change Loy, and Ziwei Liu.
\newblock Learning to prompt for vision-language models.
\newblock {\em International Journal of Computer Vision}, 130(9):2337--2348, 2022.

\bibitem[\protect\citeauthoryear{Zhou \bgroup \em et al.\egroup }{2023}]{zhou2023zegclip}
Ziqin Zhou, Yinjie Lei, Bowen Zhang, Lingqiao Liu, and Yifan Liu.
\newblock Zegclip: Towards adapting clip for zero-shot semantic segmentation.
\newblock In {\em Proceedings of the IEEE/CVF Conference on Computer Vision and Pattern Recognition}, pages 11175--11185, 2023.

\bibitem[\protect\citeauthoryear{Zou \bgroup \em et al.\egroup }{2023a}]{zou2023generalized}
Xueyan Zou, Zi-Yi Dou, Jianwei Yang, Zhe Gan, Linjie Li, Chunyuan Li, Xiyang Dai, Harkirat Behl, Jianfeng Wang, Lu~Yuan, et~al.
\newblock Generalized decoding for pixel, image, and language.
\newblock In {\em Proceedings of the IEEE/CVF Conference on Computer Vision and Pattern Recognition}, pages 15116--15127, 2023.

\bibitem[\protect\citeauthoryear{Zou \bgroup \em et al.\egroup }{2023b}]{zou2023segment}
Xueyan Zou, Jianwei Yang, Hao Zhang, Feng Li, Linjie Li, Jianfeng Gao, and Yong~Jae Lee.
\newblock Segment everything everywhere all at once.
\newblock {\em arXiv:2304.06718}, 2023.

\end{thebibliography}

\end{document}